# Spatial Coherence Loss: All Objects Matter in Salient and Camouflaged Object Detection

Ziyun Yang, *Member, IEEE*, Kevin Choy, and Sina Farsiu, *Fellow, IEEE*

*Abstract*—Generic object detection is a category-independent task that relies on accurate modeling of objectness. We show that for accurate semantic analysis, the network needs to learn all object-level predictions that appear at any stage of learning, including the pre-defined ground truth (GT) objects and the ambiguous decoy objects that the network misidentifies as foreground. Yet, most relevant models focused mainly on improving the learning of the GT objects. A few methods that consider decoy objects utilize loss functions that only focus on the single-response, i.e., the loss response of a single ambiguous pixel, and thus do not benefit from the wealth of information that an object-level ambiguity learning design can provide. Inspired by the human visual system, which first discerns the boundaries of ambiguous regions before delving into the semantic meaning, we propose a novel loss function, Spatial Coherence Loss (SCLoss), that incorporates the mutual response between adjacent pixels into the widely-used single-response loss functions. We demonstrate that the proposed SCLoss can gradually learn the ambiguous regions by detecting and emphasizing their boundaries in a self-adaptive manner. Through comprehensive experiments, we demonstrate that replacing popular loss functions with SCLoss can improve the performance of current state-of-the-art (SOTA) salient or camouflaged object detection (SOD or COD) models. We also demonstrate that combining SCLoss with other loss functions can further improve performance and result in SOTA outcomes for different applications. The codes will be released to https://github.com/TBD upon acceptance.

## I. INTRODUCTION

HUMANS can easily segment complex scenes into several parts or objects based on the semantic meaning of each object. Furthermore, they are capable of distinguishing the salient or camouflaged objects from the background based on the inherent *objectness* properties without even knowing what those objects are [1]. In computer vision, objectness represents the likelihood that each pixel is part of a foreground object. Objectness prediction, implicitly or explicitly, is a fundamental component of many computer vision tasks, and a well-trained objectness prediction model can improve downstream tasks such as weakly supervised learning, visual tracking, image retrieval, and non-photorealistic rendering [2-5].

Salient and camouflaged object detection (SOD and COD, respectively) are examples of two important tasks wherein modeling objectness plays a crucial role. The goal of SOD is to distinguish the most eye-catching objects from the background, while in COD, that goal is to find the camouflaged objects that are similar to the background [6]. Compared to other relevant tasks such as semantic segmentation [7, 8], where there exist specific classes of interests (e.g., trees, tumors, etc.), SOD and COD proceed in a category-independent manner in which they focus only on the object with specific property (e.g., saliency) rather than its actual semantic meaning [1].

Among different deep learning models trained on large-scale datasets [9, 10] for SOD and COD tasks, a major category relies on multi-scale learning to leverage the low-level features for obtaining fine-grained details [11, 12] and the high-level semantic features for better scene understanding and object locating [13, 14]. Another category exploits the pixel relationship in the latent space for improving the spatial coherence of the predictions by expanding the receptive fields [15] or using attention mechanisms or vision transformers [16-18].

We highlight two areas for improving the performance of SOD and COD methods. The first area is to exploit the distinct characteristics of the different types of errors models make and learn from in predicting objectness during training. Two major types of erroneously segmented regions during training, which we refer to as the *ambiguous regions*, are: (I) the imprecise predictions around desired ground truth (GT) regions (Fig. 1 (d)), and (II) the incorrect predictions outside the GT regions, which are usually associated with another background object, which we refer as the *decoy object* (Fig. 1 (e)). Note that these two error types reflect different problems and challenges. Specifically, the imprecise predictions around GT (such as blurred edges), reflect the network's inability to handle local detailed features, which can usually be tackled by multi-scale learning [19, 20] or edge-aware learning [21-23]. On the other hand, the decoy objects predicted by the network reflect the bottleneck of the network's semantic understanding, e.g., incorrectly assigning camouflage semantics to an object without camouflage characteristics. Due to a greater focus on

This paragraph of the first footnote will contain the date on which you submitted your paper for review, which is populated by IEEE.

This work was supported in part by the U.S. National Institutes of Health under Grants R01 EY030124 and U01 EY034687. *Corresponding author: Sina Farsiu (*e-mail: sina.farsiu@duke.edu*)*.

Ziyun Yang is with the Biomedical Engineering Department at Duke University, Durham, NC 27705, USA (e-mail: ziyun.yang@duke.edu).

Kevin Choy is with the Biomedical Engineering Department at Duke University, Durham, NC 27705, USA (e-mail: kevin.choy@duke.edu).

Sina Farsiu is with the Biomedical Engineering, Electrical and Computer Engineering, Ophthalmology, and Computer Science Departments at Duke University, Durham, NC 27705, USA (e-mail: sina.farsiu@duke.edu).

Color versions of one or more of the figures in this article are available online at http://ieeexplore.ieee.org

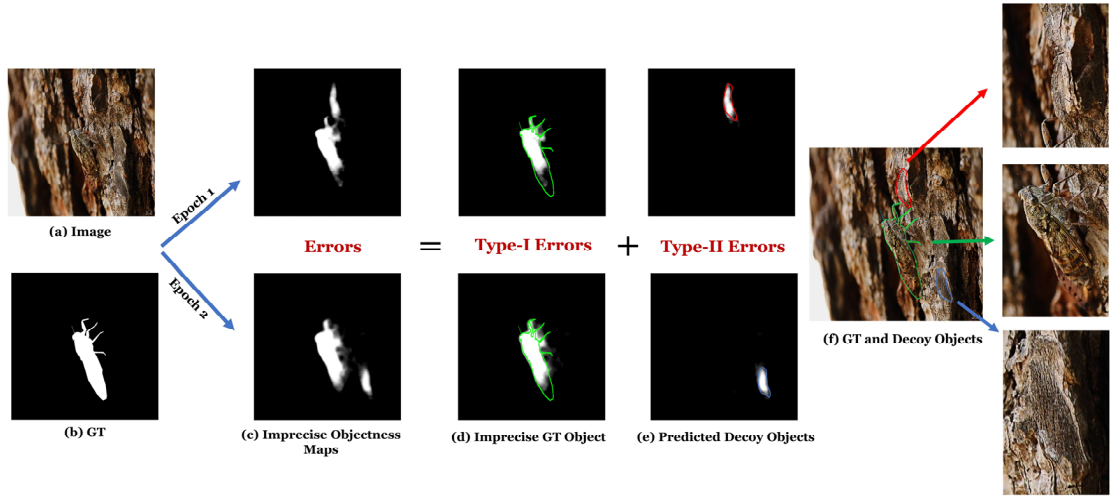

**Fig. 1**. Two categories of errors during the iterative supervised training process for segmenting an image (a) with GT mask (b) seen in the predicted objectness maps (c). Type-I error is around the predicted GT object (d). Type-II error is the incorrect predictions outside the GT regions (e), associated with background/decoy objects, such as the two tree bark sections in (f). The location of the latter category of errors across different epochs may not be bounded to a small-sized region. Building upon the previous works which mainly focused on optimizing the learning for the type-I error, to improve network's semantic understanding, we propose a loss design to self-adaptively locate the boundary of ambiguity (type-I error) and decoy objects (type-II error).

the GT, previous works [19-22] proposed learning mechanisms mainly for the first type of error, which may inadvertently overlook/weaken the learning process for decoy objects. Also, while GT objects are unambiguously located by their accompanying labels in a supervised learning scheme, the predicted decoy objects do not have a fixed spatial position. Instead, they appear in arbitrary positions that are subject to change over iterations, as in Fig. (1). Thus, related works such as boundary-aware methods [21-23] or region-aware loss designs [18, 20, 24] that focused on GT objects do *not directly applicable* to decoy objects. This highlights the need for a new targeted learning strategy for both ambiguous region types during training, to help improve the network's semantic understanding of the scene.

The second area we explore to improve the performance of object detection methods is designing a novel loss function. Most SOD and COD approaches rely on the single-response category of loss functions, with binary cross entropy (BCE) being one of the most prominent examples [25]. Single-response loss functions calculate the loss individually for each pixel without considering the loss responses of surrounding pixels. Also, when calculating the overall loss, they weight the loss calculated in each pixel equally, which dilutes the effect of important pixels (e.g., edge pixels) or the foreground pixels when the image is background-dominant. Due to these inherent shortcomings, this loss function group is not necessarily the optimal option for SOD and COD tasks [20, 26, 27]. Previous works tackled the latter limitation by finding the ambiguous pixels, i.e., the foreground/background pixels that are hard to classify correctly, either based on the single-responses [5, 28, 29] or spatial position [27, 30] and giving larger weights to the more ambiguous pixels. Unfortunately, this approach unintentionally magnifies the former limitation: since the pixels are still considered individually, wrongly labeled pixels and outlier predictions might be overly weighted, negatively affecting the training stability [29].

To take inter-pixel-responses into account, another class of methodologies [8, 31, 32] converted the objectness prediction task into several inter-pixel connectivity prediction subtasks. Of note, to improve the spatial coherence of the predicted objectness maps, i.e., how consistent the predictions are for neighboring pixels that share similar spatial features, [32] proposed to calculate the bilateral probability (mutual response) of two adjacent pixels by multiplying their unilateral probabilities (single response) that describe the likelihood of one pixel belonging to the same class as the other adjacent pixel. However, these methods required changing network architecture. They also used the acquired mutual response term as the auxiliary output of the network without further exploring its potential to establish loss functions. Meanwhile, by investigating the theoretical upper bound of the connectivity-based method, [33] highlighted the prospects for further exploiting inter-pixel relationships.

Inspired by these methods and the human visual system, *which usually first discerns the boundaries of ambiguous regions before delving into the semantic meaning* [34, 35], we propose a new scheme for loss function design to self-adaptively locate and gradually learn the boundaries of ambiguous regions in the prediction by utilizing the mutual response between adjacent pixels. Specifically, we propose the spatial coherence loss (SCLoss), in which a mutual response term takes inter-pixel response into account, and a pairwise regularization term further enhances spatial coherence. By considering the co-responses of surrounding pixels to the current input image, we enable the network to autonomously distinguish the boundaries of ambiguous regions without the need for additional mask inputs and assign weights to current ambiguous pixels based on the performance of the surrounding regions. As a result, this design ensures that the network always gives higher attention to the *boundary between ambiguous and non-ambiguous regions.* These properties enable the network to tackle the ambiguous regions gradually and smoothly by consistently being aware of and learning the transition from non-ambiguous to ambiguous regions for both the GT (type-I) and decoy (type-II) objects. To sum up, the main contributions of our work are as follows:

- We study the ambiguous regions predicted during the network learning process and propose a weighting scheme to self-adaptively locate their boundaries using inter-pixel mutual response.
- We propose a generalizable loss, SCLoss, to incorporate mutual-response of adjacent pixels into existing single-response loss functions for SOD and COD.
- We conduct extensive experiments to demonstrate incorporating SCLoss improves the performance of current state-of-the-art (SOTA) SOD and COD models.
- We show that SCLoss is compatible with other well-established loss functions in these two fields [6, 20, 24, 27] when used as an add-on loss.

## II. RELATED WORK

### A. Salient Object Detection

CNN-based methods are demonstrated to outperform the early SOD methods, which used handcrafted low-level features and heuristic priors [36-38]. The multi-scale feature aggregation model is a commonly used scheme for deep learning-based SOD. DSS [39] aggregated the multi-scale features using short connections between layers. GCPA [40] learned the global context information at different stages of layers to establish the relationship between salient regions. MINet [20] used interactive learning to integrate features from different adjacency levels to preserve the spatial consistency of the predictions. ICON [18] used three diverse aggregations and part-whole verification. EDN [13] focused on the high-level semantic features and proposed an extreme down-sampling method to learn high-level features and recover local details.

Leveraging auxiliary labels is another direction for developing SOD models. Edge map is one of the most used secondary labels in SOD. PoolNet [15] applied a multi-level model to incorporate edge supervision into different stages. EGNet [22] progressively fused the extracted salient object features with the edge features at various resolutions. Beyond edge maps, LDF [41] decoupled the saliency mask into a body map and a detailed map and learned the two maps with an interactive flow. BiconNet [32] converted the saliency mask into a connectivity mask and proposed a bilateral connectivity learning scheme to model the inter-pixel relationship between neighboring pixels to enhance spatial coherence.

### B. Camouflaged Object Detection

As with SOD, current SOTA methods for COD are CNN-based. Inspired by the bio-related nature of COD, several successful methods mimic the human visual system properties. SINet [9] designed a bio-inspired network with a searching module and a locating module to detect the concealed targets. SLSR [42] modeled the specific part of camouflaged targets to simultaneously localize, segment, and rank concealed objects. PFNet [43] simulated human identification with the distraction mining strategy. SegMaR [44] utilized a multi-scale learning scheme to integrate the segmentation, magnification, and reiteration in a coarse-to-fine manner. ZoomNet [6] mimicked human behavior when observing a vague image by zooming in and out the images via a multi-scale triplet network.

The edge labels are also widely used. BGNet [21] guided the representation learning with edge semantics. BSANet [45] utilized two-stream separated attention modules to highlight the edges of concealed objects. MGL [46] decoupled an image into two task-specific maps for locating the camouflaged objects and extracting their detailed edges. FEDER [23] decomposed the object features into different frequency bands using learnable wavelets and combined them with an auxiliary edge reconstruction flow to enhance detection. Other impactful approaches for COD include frequency domain learning [47], texture-aware learning [48], uncertainty-guided learning [49, 50], prompt learning [51] with multi-modal data, and transformer-based models [52].

### C. Loss Functions in SOD and COD

Efficient loss function design is a crucial component of constructing successful SOD and COD models. One potential solution to solve the inherent limitation of BCE loss used in many popular CNN-based methods [15, 22, 40], is to define ambiguous pixels depending on their location or loss response and weight them. Lin *et al.* [28] weighted each pixel dynamically based on the pixels' single-response and gave the ambiguous pixels higher weights. Wei *et al.* [27] assumed that the cluttered or elongated areas are prone to wrong predictions and gave higher weights to these pixels. Zhou *et al.* [30] gave higher weights to pixels near object boundaries. For COD applications, Pang *et al.* [6] proposed the UAL loss function to reduce individual pixels' uncertainty by forcing the objectness probability away from 0.5. Yet, these methods either only relied on the single-response of the prediction [6, 28] or the spatial location of the objects [27, 30], enabling the network to resolve only the pixel-level ambiguity but not the object-level ambiguity.

Another set of loss functions is the region-based category, in which a group of pixels is considered for each loss calculation. Corresponding loss functions include [18, 24] that maximized the intersection-over-union (IOU) [53] with a soft region-based loss. Zhao *et al.* [26] proposed a soft F-measure loss to maximize the relaxed F-measurement of SOD models. Pang *et al.* [20] proposed a consistency-enhanced loss (CEL) that maximized a Sørensen–Dice [54] like function. MENet [24] proposed an SSIM-based [55] loss function that computed the standard deviation and covariance for pixels and regions of the predicted map. However, these region-aware losses need object labels/masks as inputs to locate certain "regions of interest". This requirement limits their application only to GT objects, rendering them not applicable for decoy objects that do not have fixed locations/masks during network training.

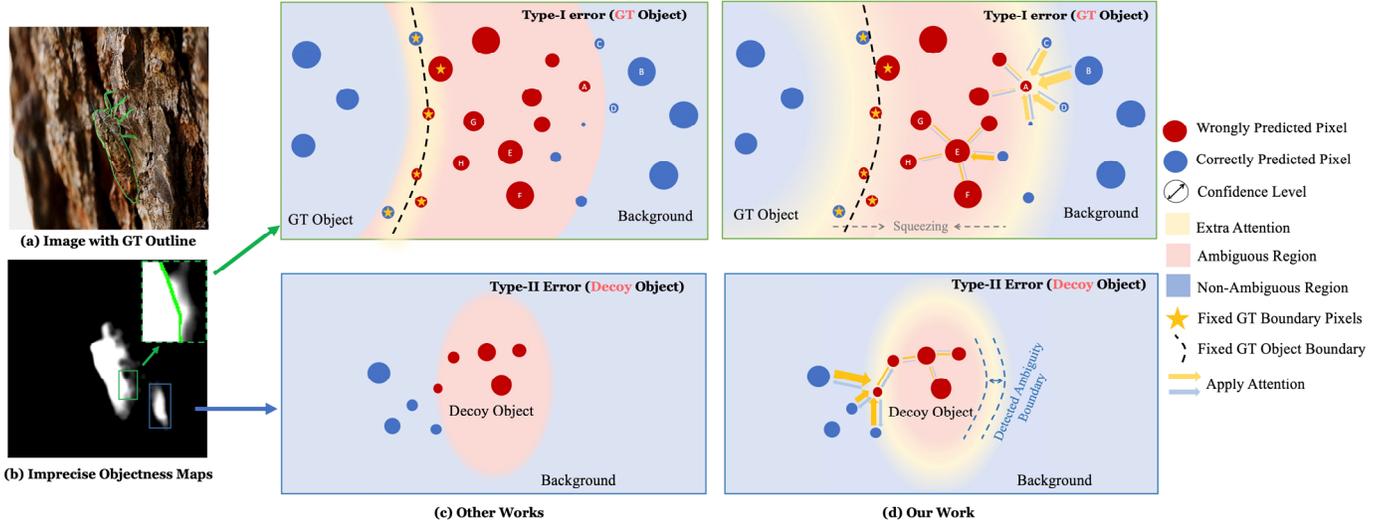

**Fig. 2.** Schematic representation of loss function interaction with two types of ambiguous regions: around the GT object (type-I error) and the decoy object (type-II error). The predicted objectness map of the image in (a) is shown in (b) which expected to include imprecise predictions at an arbitrary epoch. In previous works [21, 23, 27, 30], extra attention is applied only to GT regions (e.g., edges) near the fixed GT boundary pixels (c, top). No extra attention is applied for other ambiguous pixels that are not near GT objects and belong to a decoy object (c, bottom). In (d), we identify and emphasize boundaries between ambiguous and non-ambiguous regions by computing mutual responses between neighboring pixels, in both types of error regions. Specifically, one pixel receives attention from all its adjacent pixels. The attention (scaled by the arrow thickness) increases with the confidence level (scaled by the dot size) of the surrounding correctly predicted pixels, allowing pixels at ambiguous regions' boundaries to receive the highest attention. E.g., for pixel A, neighboring pixels B, C, and D that are correctly predicted will increase the contribution to the loss by pixel A. At the same time, pixels with higher ambiguity, like pixel E, have reduced weight due to the limited number of correctly predicted pixels in their adjacency.

Finally, we note that improved performance in current SOTA models, such as MENet [24], EDN [13], and ICON [18], was achieved in part by combining different loss functions rather than using any of them individually. Inspired by these, in Section VI (E), after demonstrating that the proposed SCLoss outperforms other SOTA losses in one-on-one comparisons, we show that it is also compatible with other loss functions and, when combined with them, results in overall best outcomes.

## III. METHODOLOGY

### A. Mutual Response as Edge-aware Ambiguity Weighting

Cross entropy is a metric to evaluate ambiguity for the objectness prediction of a single pixel [56]. However, this ambiguity is assessed only by the single pixel's response to the given label without considering the interactions between the adjacent pixels, which might or might not share the same labels. Previous works that studied pixel connectivity have elucidated that modeling the mutual response among multiple (especially neighboring) pixels enhances spatial coherence between pixels [31, 32]. Unlike traditional approaches that classify individual pixels separately, these works use the collective response of pixels to the same class label. The modeling process is formulated as:

$$arg \min_{\theta}(loss(\mathcal{F}_\theta(X)_i * \mathcal{F}_\theta(X)_j, Y_{ij})), \quad (1)$$

where $X$ and $Y$ are the image input and connectivity mask, respectively. $\mathcal{F}$ is a connectivity-based network architecture parameterized by $\theta$. The loss function $loss$ quantifies the difference between the mutual response of two neighboring pixels $i$ and $j$ and the shared connectivity label $Y_{ij}$. The modeling process in these early works required modifying the network architecture.

Inspired by these works, we incorporate the mutual response in loss design. Specifically, for two adjacent pixels with labels $n_i$ and $n_j$, and network predictions $p_i$ and $p_j$, we define the mutual response as:

$$L_{mutual}(p_i, p_j, n_i, n_j) = n_i n_j \log(p_i p_j) + (1 - n_i n_j)\log(1 - p_i p_j). \quad (2)$$

Compared to the single-response, this term combines the predictions of the surrounding pixels and is region-aware. The inverse of mutual response is a weighting term that is edge-aware at the ambiguous regions:

$$w_{mutual\_response} = \frac{1}{L_{mutual} + T}, \quad (3)$$

where T is a constant used to scale the weight at the non-ambiguous region and avoid overflow. Thus, the weight of each pixel changes according to the accuracy of its surrounding region predictions. In non-ambiguous regions, where the scale of the loss itself is small, this weighting scheme does not cause scalable differences in actual loss. However, in ambiguous regions, the boundaries between ambiguous and non-ambiguous regions receive the highest attention, allowing the network to locate the boundaries of decoy objects and gradually learn the transition from non-ambiguous to ambiguous regions. Fig. 2 shows our weighting scheme in the ambiguous regions around GT (top) and decoy (bottom) objects. Note that the proposed weighting scheme handles both types of ambiguous regions, while the existing works [21, 23, 27, 30] only focus on the former case.

### B. Proposed Spatial Coherence Loss

The proposed Spatial Coherence Loss (SCLoss) combines the classic single-response loss function with a mutual response term and a pairwise regularization term:

$$L = \frac{Single-Response}{Mutual\ Response + Pairwise\ Regularization}, \quad (4)$$

where the single-response term is the standard pixel-wise loss function (i.e., the loss response from a single pixel) of choice. The mutual response term calculates the co-response of two adjacent pixels, and the pairwise regularization term further enhances the spatial coherence in foreground areas. In brief, we penalize or encourage the objectness probability prediction of individual pixels by the performance of their surrounding pixels to enhance the spatial consistency of the final prediction. In the following, we present our proposed default choices for each of these terms in SOD and COD. In Sections VI (B) and VI (D), we discuss the utility of alternative terms, which show consistent performance improvement as compared to their corresponding baseline loss function.

### C. Instantiation

We call a pixel that is $k$ pixels apart from a given pixel as its $k^{th}$-level adjacent pixel. Then, for pixel $i$, the SCLoss is the weighted sum of sub-losses at different adjacency levels:

$$L_i = \sum_{k=1}^{K} \left(\frac{1}{2}\right)^{k-1} L_i^k, \quad (5)$$

where $K$ is the maximum adjacency level (i.e., defining a $(2K+1) \times (2K+1)$ patch). For the $k^{th}$ level, the loss $L_i^k$ for pixel $i$ is defined in the format of (4):

$$L_i^k = \frac{1}{N} \sum_{j \in k^{th}} \frac{L_{single}(p_i, n_i)}{L_{mutual}(p_i, p_j, n_i, n_j) + \alpha f(p_i, p_j)}, \quad (6)$$

where $j$ indexes pixels which are the $k^{th}$-level adjacents of pixel $i$, $p_i$ and $p_j$ are the predicted objectness probability at each pixel, and $n_i$ and $n_j$ are the labels. $N$ is the total number of the $k^{th}$-level adjacent pixels of $i$. $L_{single}$ is the single-response term, $L_{mutual}$ is the mutual response term, and $f$ is the pairwise regularization term. Scalar $\alpha$ balances the contributions of the mutual response and pairwise regularization terms (See Section VI (C)).

**Single-response term.** Our default choice for SOD and COD tasks is the BCE loss:

$$L_{single}(p_i, n_i) = n_i \log(p_i) + (1 - n_i)\log(1 - p_i), \quad (7)$$

where $p_i$ is the predicted objectness probability, and $n_i$ is the label of pixel $i$. Other possible choices for this term are discussed in Section VI (B).

**Mutual response term.** Equation (2) defines this term, which incorporates both the inter-pixel relation and pixel location information into the overall loss calculation. In this setting, the determinacies of the surrounding pixels' predictions play an important role in weighting the current pixel's loss calculation. When the two adjacent pixels belong to the same class, the weight of one pixel increases as the other pixel gets a more accurate prediction. When the two pixels are not from the same class, i.e., at the object edges, a foreground edge pixel's weight will be increased when: 1) it is wrongly predicted or 2) the other pixel gets correctly predicted. There are two benefits to this: first, the surrounding pixels encourage the network to make a high spatially coherent prediction; second, when there are ambiguous regions, the mutual response term gives higher weights to the boundary of the ambiguous region, which is surrounded by correctly predicted pixels, while giving smaller weights to the region's center to encourage the network to gradually learn the transition from non-ambiguous region to ambiguous regions.

**Pairwise regularization term**. This term further enhances the spatial coherence of the foreground region. Among many possible options studied in Section VI (D), our default choice is the Gaussian function [57, 58]:

$$f(p_i, p_j) = e^{-p_i p_j}, \quad (8)$$

which locally adjusts the weights by the dot product of the predicted objectness probability. We give the highest weights to the current pixel when both pixels in the pair are predicted as foreground to 1) ensure a high spatial coherence in the foreground prediction; 2) minimize the false positive detections.

## IV. SALIENT OBJECT DETECTION

### A. Dataset and Evaluation Metrics

**Dataset:** We used the training part of DUTS [10] dataset, DUT-TR, for training. For the testing, all methods were evaluated on the following five popular SOD benchmark datasets. ECSSD [59] has 1,000 images in which the foreground objects are semantically meaningful, and both foregrounds and backgrounds are structurally complex. HKU-IS [19] has 4,447 images, most containing multiple scattered salient objects. DUT-OMRON [3] has 5,168 images with relatively complex backgrounds. PASCAL-S [60] is a subset of PASCAL-VOC 2010, containing 850 images. DUTS-TE [10] is the testing part of the DUTS dataset, with 5,019 images.

**Evaluation Criteria:** We assessed SOD models with the Mean Absolute Error (MAE) [37], the F-measure [61], including the adaptive mean ($F_\beta^{adp}$), weighted ($F_\beta^w$) F-measure [62], the

TABLE I
QUANTITATIVE COMPARISON OF SOD MODELS ON PASCAL, DUT-TE, HKU-IS, OMRON, AND ECSSD DATASETS, WITHOUT AND WITH SCLOSS AS AN ADD-ON TO EACH MODEL'S RECOMMENDED LOSS FUNCTION. THE BEST RESULTS FOR EACH METRIC AND DATASET ARE IN **BOLD**.

| Model | Year | | PASCAL (850 images) | | | | | DUT-TE (5019 images) | | | | | HKU-IS (4447 images) | | | | | OMRON (5168 images) | | | | | ECSSD (1000 images) | | | | |
|---|---|---|---|---|---|---|---|---|---|---|---|---|---|---|---|---|---|---|---|---|---|---|---|---|---|---|---|
| | | | $F_\beta^{adp}$ | MAE | $E_m$ | $F_\beta^w$ | $S_m$ | $F_\beta^{adp}$ | MAE | $E_m$ | $F_\beta^w$ | $S_m$ | $F_\beta^{adp}$ | MAE | $E_m$ | $F_\beta^w$ | $S_m$ | $F_\beta^{adp}$ | MAE | $E_m$ | $F_\beta^w$ | $S_m$ | $F_\beta^{adp}$ | MAE | $E_m$ | $F_\beta^w$ | $S_m$ |
| GCPA [40] | 2020 | | .833 | .062 | .849 | .820 | .862 | .814 | .039 | .888 | .819 | .890 | .901 | .030 | .950 | .891 | .921 | .731 | .061 | .849 | .716 | .829 | .919 | .035 | .921 | .904 | .927 |
| + SCLoss | | | .845 | .059 | .858 | .827 | .860 | .844 | .035 | .903 | .835 | .891 | .920 | .028 | .955 | .901 | .921 | .768 | .056 | .867 | .747 | .837 | .933 | .032 | .927 | .916 | .929 |
| MINet [20] | 2020 | | .842 | .067 | .865 | .822 | .857 | .837 | .035 | .902 | .840 | .893 | .916 | .026 | .956 | .908 | .926 | .763 | .053 | .870 | .751 | .842 | .926 | .035 | .924 | .911 | .925 |
| + SCLoss | | | .849 | .065 | .867 | .827 | .858 | .858 | .034 | .912 | .847 | .890 | .923 | .025 | .958 | .912 | .924 | .778 | .051 | .876 | .756 | .839 | .935 | .032 | .928 | .918 | .924 |
| ITSD [30] | 2020 | CNN-Based Backbone | .812 | .069 | .858 | .817 | .856 | .806 | .041 | .891 | .818 | .884 | .901 | .029 | .953 | .896 | .920 | .753 | .058 | .862 | .751 | .843 | .903 | .034 | .925 | .910 | .927 |
| + SCLoss | | | .834 | .069 | .858 | .822 | .854 | .829 | .038 | .901 | .830 | .885 | .911 | .029 | .954 | .899 | .918 | .769 | .053 | .871 | .758 | .844 | .924 | .033 | .930 | .914 | .925 |
| EDN [13] | 2022 | | .841 | .064 | .856 | .827 | .858 | .842 | .036 | .900 | .838 | .888 | .912 | .027 | .954 | .905 | .922 | .774 | .050 | .868 | .764 | .840 | .927 | .033 | .924 | .918 | .925 |
| + SCLoss | | | .852 | .062 | .860 | .833 | .859 | .851 | .036 | .905 | .843 | .887 | .920 | .027 | .955 | .907 | .921 | .781 | .048 | .872 | .767 | .840 | .933 | .031 | .925 | .920 | .927 |
| MENet [24] | 2023 | | .861 | .057 | .866 | .842 | .867 | .871 | .028 | .919 | .868 | .904 | .925 | .024 | .958 | .916 | .928 | .788 | .047 | .880 | .771 | .850 | .933 | .031 | .923 | .920 | .928 |
| + SCLoss | | | .868 | .057 | .867 | .848 | .870 | .877 | .028 | .921 | .875 | .909 | .933 | .022 | .964 | .926 | .933 | .796 | .047 | .883 | .784 | .856 | .942 | .026 | .933 | .932 | .937 |
| ICON-R [18] | 2022 | | .842 | .065 | .860 | .826 | .860 | .837 | .037 | .899 | .835 | .887 | .909 | .029 | .949 | .899 | .919 | .769 | .057 | .869 | .759 | .844 | .928 | .033 | .929 | .917 | .929 |
| + SCLoss | | | .851 | .064 | .863 | .831 | .861 | .852 | .036 | .907 | .843 | .892 | .921 | .027 | .955 | .911 | .923 | .783 | .055 | .875 | .768 | .846 | .937 | .029 | .933 | .926 | .932 |
| ICON-P [18] | 2022 | Transformer | .865 | .052 | .872 | .849 | .881 | .868 | .025 | .919 | .880 | .917 | .921 | .023 | .961 | .922 | .934 | .794 | .047 | .885 | .791 | .864 | .934 | .024 | .928 | .932 | .939 |
| + SCLoss | | | **.883** | **.048** | **.883** | **.870** | **.888** | **.891** | **.024** | **.930** | **.894** | **.923** | **.936** | **.020** | **.966** | **.932** | **.940** | **.808** | **.045** | **.893** | **.804** | **.870** | **.946** | **.022** | **.934** | **.939** | **.944** |

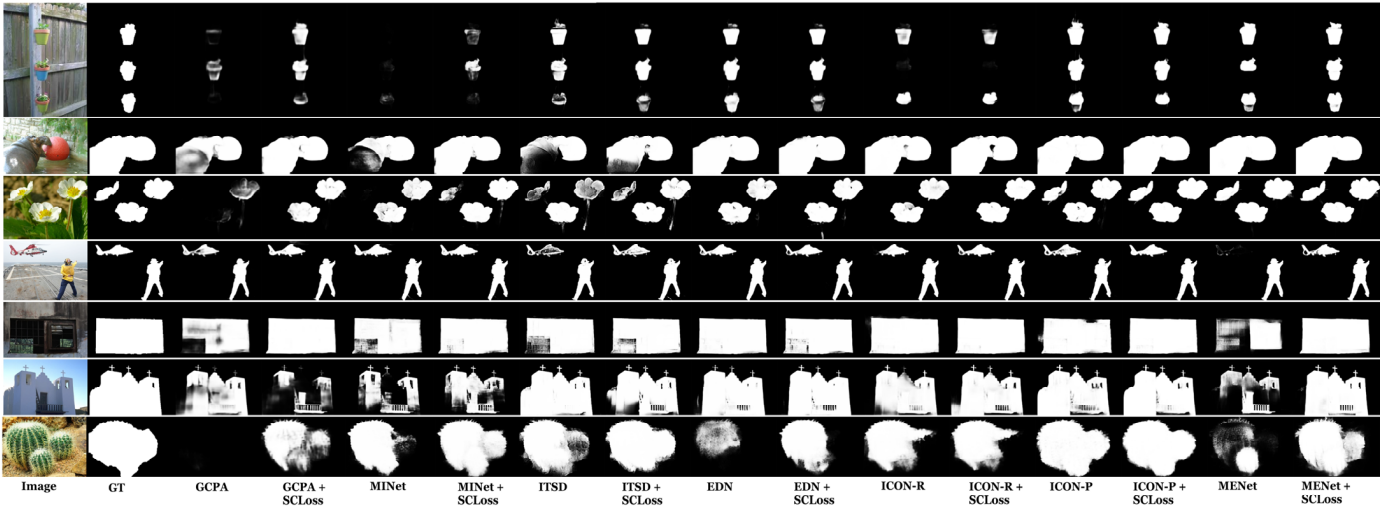

**Fig. 3.** Qualitative comparison between different SOD models trained with/without SCLoss as an add-on to each model's recommended loss function.

Enhanced-alignment Measure ($E_m$) [63], and the S-measure ($S_m$) [64].

### B. Implementation Details

Our experiments were based on seven state-of-the-art SOD models: GCPA [40], MINet [20], ITSD [30], EDN [13], ICON [18] (including ICON-R and ICON-P), and MENet [24]. Among these models, ICON-P used a transformer-based backbone PVTv2 [65] while others used CNN-based backbones. For all models, we added SCLoss to their original proposal and followed the suggested implementation details in corresponding papers to train both the baselines and the SCLoss incorporated version. All the experiments were done on PyTorch 1.7.0 [66] and were performed with an NVIDIA GeForce GTX 3090 Ti GPU.

### C. Experimental Results

**Quantitative comparison**: Table I compares each baseline with a version trained with SCLoss as a simple add-on loss function. The proposed SCLoss version consistently improved all models across all six utilized metrics. Among these metrics, the $F_\beta^{adp}$ improved most, which suggests that our loss can uniformly highlight the salient object region, implying the highest spatial coherence [32]. Note that adding SCLoss to the most recent SOTA models: MENet [24], ICON-R (ResNet as the backbone), and ICON-P (vision transformer PVTv2 as the backbone) [18], improved their performance for almost all metrics.

TABLE II
QUANTITATIVE COMPARISON OF COD MODELS ON CAMO, CHAMELEON, COD10K, AND NC4K DATASETS, WITHOUT AND WITH SCLOSS AS AN ADD-ON TO EACH MODEL'S RECOMMENDED LOSS FUNCTION. THE BEST RESULTS IN EACH METRIC AND DATASET ARE IN **BOLD**.

| Model | Year | | CHAMELEON (67 images) | | | | | | CAMO (250 images) | | | | | | COD10K (2026 images) | | | | | | NC4K (4121 images) | | | | | |
|---|---|---|---|---|---|---|---|---|---|---|---|---|---|---|---|---|---|---|---|---|---|---|---|---|---|---|
| | | | MAE | $F_\beta^{adp}$ | $F_\beta^w$ | $E_\varphi^{adp}$ | $E_\varphi^m$ | $S_m$ | MAE | $F_\beta^{adp}$ | $F_\beta^w$ | $E_\varphi^{adp}$ | $E_\varphi^m$ | $S_m$ | MAE | $F_\beta^{adp}$ | $F_\beta^w$ | $E_\varphi^{adp}$ | $E_\varphi^m$ | $S_m$ | MAE | $F_\beta^{adp}$ | $F_\beta^w$ | $E_\varphi^{adp}$ | $E_\varphi^m$ | $S_m$ |
| GCPA [40] | 2020 | | .049 | .757 | .722 | .881 | .866 | .863 | .089 | .721 | .651 | .848 | .805 | .787 | .046 | .590 | .581 | .788 | .816 | .792 | .058 | .742 | .703 | .866 | .854 | .830 |
| + **SCLoss** | | | .035 | .799 | .766 | .911 | .900 | .871 | .086 | .741 | .660 | .848 | .800 | .782 | .041 | .641 | .612 | .829 | .825 | .786 | .054 | .774 | .721 | .884 | .859 | .825 |
| SLSR [42] | 2021 | CNN-Based Backbone | .033 | .811 | .802 | .924 | .933 | .881 | .084 | .759 | .691 | .863 | .836 | .785 | .039 | .675 | .655 | .860 | .869 | .802 | .054 | .778 | .739 | .891 | .881 | .831 |
| + **SCLoss** | | | .032 | .823 | .810 | .932 | .935 | .882 | .080 | .764 | .701 | .867 | .850 | .795 | .037 | .686 | .667 | .871 | .881 | .806 | .051 | .784 | .748 | .895 | .889 | .833 |
| SINet-v2 [69] | 2022 | | .030 | .816 | .816 | .929 | .941 | .888 | .070 | .779 | .743 | .885 | .882 | .820 | .037 | .682 | .680 | .864 | .887 | .815 | .048 | .792 | .769 | .901 | .902 | .847 |
| + **SCLoss** | | | .028 | .829 | .823 | .937 | .943 | .891 | .069 | .789 | .749 | .890 | .883 | .823 | .035 | .696 | .688 | .875 | .891 | .817 | .047 | .799 | .772 | .905 | .902 | .847 |
| BGNet [21] | 2022 | | .028 | .838 | .840 | .934 | .939 | .896 | .075 | .778 | .736 | .860 | .852 | .803 | .032 | .738 | .719 | .901 | .897 | .830 | .045 | .813 | .786 | .909 | .901 | .851 |
| + **SCLoss** | | | .028 | .846 | .847 | .939 | .948 | .898 | .069 | .791 | .753 | .879 | .875 | .815 | .031 | .740 | .721 | .901 | .898 | .831 | .044 | .817 | .790 | .911 | .905 | .852 |
| FEDER [23] | 2023 | | .029 | .848 | .836 | .946 | .949 | .887 | .076 | .768 | .715 | .862 | .850 | .785 | .033 | .732 | .707 | .901 | .896 | .817 | .046 | .812 | .777 | .907 | .899 | .841 |
| + **SCLoss** | | | .027 | .848 | .840 | .947 | **.951** | .890 | .072 | .784 | .737 | .877 | .869 | .803 | .032 | .735 | .715 | .898 | .901 | .824 | .044 | .820 | .789 | .914 | .909 | .849 |
| FSPNet [52] | 2023 | Transformer | .025 | .845 | .839 | .944 | .938 | .904 | .051 | .828 | .799 | .919 | .899 | .856 | .027 | .729 | .732 | .896 | .894 | **.851** | .035 | .823 | .816 | .922 | .915 | **.878** |
| + **SCLoss** | | | **.022** | **.873** | **.855** | **.962** | .946 | **.904** | **.049** | **.836** | **.810** | **.921** | **.908** | **.858** | **.026** | **.753** | **.742** | **.908** | **.899** | .847 | **.035** | **.836** | **.823** | **.926** | **.920** | .877 |

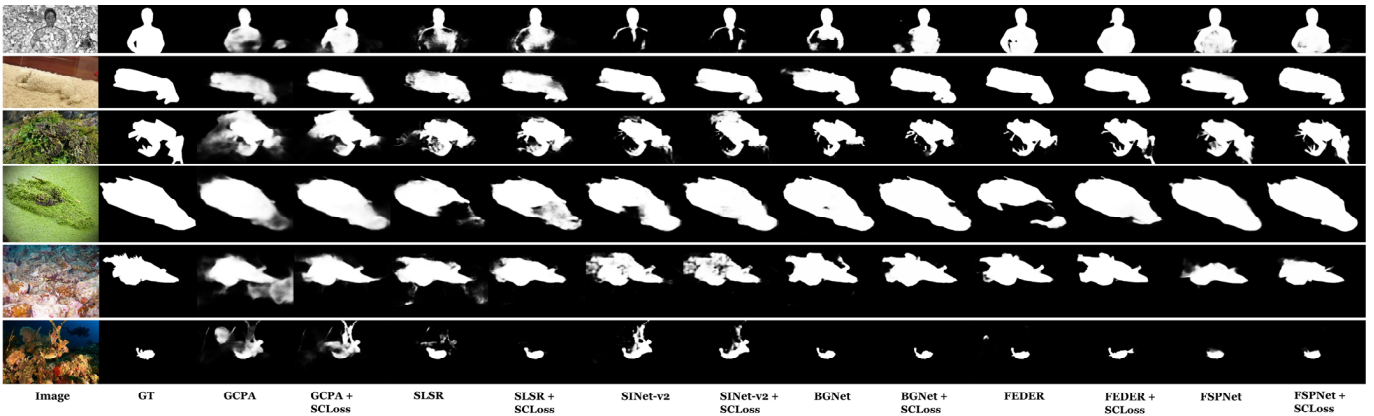

**Fig. 4.** Comparison between different COD models trained with/without SCLoss as an add-on to each model's recommended loss function.

**Qualitative comparison**: Fig. 3 demonstrates the impact of adding SCLoss to different models for different scenarios, including detecting small objects, multiple instances, high-contrast objects, and low-contrast object backgrounds. In all challenging cases, adding SCLoss resulted in visually better object detection compared to the corresponding baseline, where the predicted saliency maps are uniformly highlighted, and the boundaries of the targets are more precise and complete.

V. CAMOUFLAGED OBJECT DETECTION

*A. Dataset and Evaluation Metrics*

**Dataset:** We followed the popular setting in [9] and used the mix of the training part of CAMO [67] and COD10K [9] as the training dataset. For the testing dataset, we used four popular COD datasets: CAMO [67], CHAMELEON [68], COD10K [9], and NC4K [42]. CAMO is a large-scale COD dataset, containing 1,250 camouflaged images and 1,250 non-camouflaged images. CHAMELEON is a smaller dataset that only contains 76 hand-annotated images collected online. COD10K is currently the largest COD dataset, which contains 5,066 camouflaged, 3,000 background, and 1,934 non-camouflaged images. NC4K is another recently released large-scale COD testing dataset that contains 4,121 images.

TABLE III
ABLATION STUDY ON THE ADJACENCY LEVELS FOR SCLOSS IN SOD (TOP) AND COD (DOWN). THE TRAINING GPU MEMORY COSTS ON ARE ALSO REPORTED.

| | SOD - ECSSD | | | | | SOD - HKU-IS | | | | | SOD - OMRON | | | | | SOD - PASCAL | | | | | SOD - DUT-TE | | | | |
|---|---|---|---|---|---|---|---|---|---|---|---|---|---|---|---|---|---|---|---|---|---|---|---|---|---|
| Exp. | $F_\beta^{adp}$ | MAE | $F_\beta^w$ | $S_m$ | $E_m$ | $F_\beta^{adp}$ | MAE | $F_\beta^w$ | $S_m$ | $E_m$ | $F_\beta^{adp}$ | MAE | $F_\beta^w$ | $S_m$ | $E_m$ | $F_\beta^{adp}$ | MAE | $F_\beta^w$ | $S_m$ | $E_m$ | $F_\beta^{adp}$ | MAE | $F_\beta^w$ | $S_m$ | $E_m$ |
| GCPA | .919 | .035 | .904 | .927 | .921 | .901 | .030 | .891 | .921 | .950 | .731 | .061 | .716 | .829 | .849 | .833 | .062 | .820 | .862 | .849 | .814 | .039 | .819 | .890 | .888 |
| K=1 | .930 | .033 | .912 | .927 | .927 | .912 | .030 | .897 | .919 | .952 | .754 | .055 | .732 | .830 | .866 | .842 | .060 | .824 | .859 | .854 | .837 | .036 | .829 | .888 | .903 |
| K=2 | .933 | .032 | .916 | .929 | .927 | .920 | .028 | .901 | .921 | .955 | .768 | .056 | .747 | .837 | .867 | .845 | .059 | .827 | .860 | .858 | .844 | .035 | .835 | .891 | .903 |
| K=3 | .930 | .033 | .914 | .927 | .927 | .916 | .029 | .898 | .918 | .954 | .760 | .055 | .740 | .832 | .866 | .845 | .059 | .824 | .857 | .864 | .846 | .035 | .834 | .889 | .907 |

| | COD - CAMO | | | | | | COD - CHAMELEON | | | | | | COD - COD10K | | | | | | COD - NC4K | | | | | | Training Memory (GB) |
|---|---|---|---|---|---|---|---|---|---|---|---|---|---|---|---|---|---|---|---|---|---|---|---|---|---|
| Exp. | $S_m$ | $F_\beta^w$ | MAE | $E_\varphi^{adp}$ | $E_\varphi^m$ | $F_\beta^{adp}$ | $S_m$ | $F_\beta^w$ | MAE | $E_\varphi^{adp}$ | $E_\varphi^m$ | $F_\beta^{adp}$ | $S_m$ | $F_\beta^w$ | MAE | $E_\varphi^{adp}$ | $E_\varphi^m$ | $F_\beta^{adp}$ | $S_m$ | $F_\beta^w$ | MAE | $E_\varphi^{adp}$ | $E_\varphi^m$ | $F_\beta^{adp}$ | |
| SLSR | .785 | .691 | .084 | .863 | .836 | .759 | .881 | .802 | .033 | .924 | .933 | .811 | .802 | .655 | .039 | .860 | .869 | .675 | .831 | .739 | .054 | .891 | .881 | .778 | 7.6 |
| K=1 | .787 | .696 | .083 | .863 | .838 | .759 | .882 | .804 | .033 | .928 | .935 | .812 | .805 | .662 | .038 | .864 | .879 | .677 | .833 | .746 | .052 | .894 | .887 | .780 | 9.3 |
| K=2 | .795 | .701 | .080 | .867 | .850 | .764 | .882 | .810 | .032 | .932 | .935 | .823 | .806 | .667 | .037 | .871 | .881 | .686 | .833 | .748 | .051 | .895 | .889 | .784 | 11.2 |
| K=3 | .790 | .700 | .082 | .865 | .844 | .760 | .880 | .807 | .033 | .931 | .934 | .823 | .805 | .668 | .037 | .871 | .881 | .688 | .832 | .748 | .052 | .895 | .888 | .785 | 13.0 |

TABLE IV
ABLATION STUDY ON SELECTION OF $L_{single}$. WE USED TWO OTHER COMMON SINGLE-RESPONSE BASED LOSSES, MSE AND L1 LOSS AS OUR BASELINES AND ESTABLISHED SCLOSS ON EACH OF THEM FOR SOD AND COD EXPERIMENTS.

| | SOD - GCPA - ECSSD | | | | SOD - GCPA - HKU-IS | | | | COD – SLSR - CAMO | | | | | | COD – SLSR - NC4K | | | | | |
|---|---|---|---|---|---|---|---|---|---|---|---|---|---|---|---|---|---|---|---|---|
| Loss | $F_\beta^{adp}$ | MAE | $S_m$ | $E_m$ | $F_\beta^{adp}$ | MAE | $S_m$ | $E_m$ | $S_m$ | $F_\beta^w$ | MAE | $E_\varphi^{adp}$ | $E_\varphi^m$ | $F_\beta^{adp}$ | $S_m$ | $F_\beta^w$ | MAE | $E_\varphi^{adp}$ | $E_\varphi^m$ | $F_\beta^{adp}$ |
| L1 Loss | .924 | .033 | .915 | .918 | .912 | .028 | .910 | .950 | .753 | .666 | .084 | .836 | .814 | .741 | .796 | .718 | .055 | .879 | .859 | .779 |
| L1 SCLoss | .931 | .032 | .915 | .920 | .919 | .027 | .907 | .947 | .775 | .711 | .078 | .859 | .854 | .763 | .810 | .744 | .052 | .889 | .885 | .785 |
| MSE Loss | .882 | .044 | .922 | .919 | .887 | .039 | .917 | .945 | .785 | .645 | .090 | .851 | .798 | .728 | .817 | .680 | .063 | .860 | .835 | .735 |
| MSE SCLoss | .924 | .037 | .923 | .924 | .911 | .032 | .917 | .952 | .784 | .677 | .080 | .876 | .821 | .765 | .816 | .713 | .056 | .883 | .856 | .772 |

**Evaluation Criteria:** We used six evaluation metrics, including MAE, adaptive F-measure ($F_\beta^{adp}$), weighted F-measure ($F_\beta^w$) [62], adaptive E-measure ($E_\beta^{adp}$) [63], mean E-measure ($E_\varphi^m$), and S-measure ($S_m$) [64].

### B. Implementation Details

Our experiments were based on six state-of-the-art SOD models: GCPA [40], SLSR [42], BGNet [21], SINet-v2 [69], FEDER [23], and FSPNet [52]. FSPNet used a transformer-based backbone, while others used CNN-based backbones. We followed the suggested implementation details in the corresponding papers, and for the sake of fairness, we used the same settings for retraining SCLoss. All the experiments were done on PyTorch 1.7.0 [66] and were performed with an NVIDIA GeForce GTX 3090 Ti GPU.

### C. Experimental Results

**Quantitative comparison:** Table II shows the impact of adding SCLoss to the baseline models. SCLoss improved the performance of all models across all metrics. This includes improving the SOTA results from the most recent models, FEDER [23] and FSPNet [52].

**Qualitative comparison:** Fig. 4 showcases a few challenging scenes for comparison, including camouflaged objects of different sizes and backgrounds. For each model, adding SCLoss resulted in visually more acceptable detection compared to the baseline. Particularly, our results have fewer false positive predictions, higher confidence, and more precise predictions in the foreground objects.

## VI. ABLATION STUDY

To study the impact of each SCLoss component, we used two widely used baselines GCPA [33] and SLSR [36], for the SOD and COD experiments, respectively.

### A. Adjacency Level of SCLoss

We first study the impact of the maximum adjacency level K in (5) on SCLoss. Table III shows the improvements gained by adding SCLoss with the total level K of 1, 2, and 3 to baseline models. The GPU-memory cost as a function of K in the training phase is listed in the last column of Table III. Comparison of the baseline with the case of K=1, shows that incorporating even the simplest form of the SCLoss into the training greatly improves the baseline performance. Increasing the adjacency level (K=2), further improves the performance, but performance improvements are saturated at higher K values (which also have slightly higher memory costs in the training process). Thus, we set K=2 as our default choice for the best balance of performance and GPU-memory cost.

### B. Selection of Single-response Term

The choice for $L_{single}$ depends on the task at hand. Table IV shows the impact of using two (alternative to BCE) widely used single-response loss functions: MSE and L1 loss. We compared the performance of GCPA for SOD and SLSR for COD tasks, when using MSE or L1 Loss alone or as the $L_{single}$ in SCLoss as the loss function. The consistent improvements gained by using the SCLoss version of these loss functions suggest that other single-response terms may also benefit from our proposed loss scheme in (4).

TABLE V
ABLATION STUDY ON SELECTION OF $\alpha$. DUE TO LIMITED SPACE, WE USED ONE DATASET FROM EACH TASK TO COMPARE THE IMPACT OF $\alpha = 0.5, 1,$ AND $2$.

| | SOD - GCPA - ECSSD | | | | | COD - SLSR - CAMO | | | | | |
|---|---|---|---|---|---|---|---|---|---|---|---|
| Exp. | $F_\beta^{adp}$ | MAE | $F_\beta^w$ | $S_m$ | $E_m$ | $S_m$ | $F_\beta^w$ | MAE | $E_\varphi^{adp}$ | $E_\varphi^m$ | $F_\beta^{adp}$ |
| $\alpha=0.5$ | .933 | .032 | .913 | .926 | .926 | .792 | .704 | .083 | .861 | .848 | .759 |
| $\alpha=1$ | .933 | .032 | .916 | .929 | .927 | .795 | .701 | .080 | .867 | .850 | .764 |
| $\alpha=2$ | .928 | .035 | .908 | .925 | .923 | .785 | .693 | .084 | .864 | .836 | .758 |

TABLE VI
ABLATION STUDY ON THE CHOICE FOR PAIRWISE REGULARIZATION TERM.

| | SOD - GCPA - ECSSD | | | | | COD - SLSR - CHAMELEON | | | | | |
|---|---|---|---|---|---|---|---|---|---|---|---|
| Exp. | $F_\beta^{adp}$ | MAE | $F_\beta^w$ | $S_m$ | $E_m$ | $S_m$ | $F_\beta^w$ | MAE | $E_\varphi^{adp}$ | $E_\varphi^m$ | $F_\beta^{adp}$ |
| Constant | .929 | .033 | .912 | .927 | .922 | .876 | .802 | .033 | .929 | .925 | .817 |
| Distance | .931 | .033 | .916 | .928 | .926 | .883 | .805 | .033 | .928 | .934 | .814 |
| Gaussian | .933 | .032 | .916 | .929 | .927 | .882 | .810 | .032 | .932 | .935 | .823 |

TABLE VII
COMPARISON BETWEEN SCLOSS AND OTHER LOSS FUNCTIONS IN SOD AND COD TASKS. GCPA WAS USED FOR SOD AND SLSR WAS USED FOR COD.

| | SOD - GCPA - ECSSD | | | | SOD - GCPA - PASCAL | | | | SOD - GCPA - DUT-TE | | | | COD – SLSR - CAMO | | | | | | | COD – SLSR - NC4K | | | | | | |
|---|---|---|---|---|---|---|---|---|---|---|---|---|---|---|---|---|---|---|---|---|---|---|---|---|---|---|
| Loss | $F_\beta^{adp}$ | MAE | $S_m$ | $E_m$ | $F_\beta^{adp}$ | MAE | $S_m$ | $E_m$ | $F_\beta^{adp}$ | MAE | $S_m$ | $E_m$ | $S_m$ | $F_\beta^w$ | MAE | $E_\varphi^{adp}$ | $E_\varphi^m$ | $E_\varphi^x$ | $F_\beta^{adp}$ | $S_m$ | $F_\beta^w$ | MAE | $E_\varphi^{adp}$ | $E_\varphi^m$ | $E_\varphi^x$ | $F_\beta^{adp}$ |
| BCE | .919 | .035 | .927 | .921 | .833 | .062 | .862 | 849 | .814 | .039 | .890 | .888 | .788 | .649 | .089 | .857 | .797 | .863 | .737 | .821 | .684 | .061 | .867 | .835 | .897 | .745 |
| CEL [20] | .923 | .033 | .926 | .921 | .836 | .063 | .856 | .846 | .829 | .037 | .889 | .896 | .787 | .684 | .082 | .870 | .832 | .870 | .752 | .822 | .722 | .055 | .885 | .866 | .896 | .768 |
| Structure [27] | .926 | .032 | .927 | .923 | .841 | .061 | .858 | .848 | .841 | .036 | .892 | .902 | .785 | .691 | .084 | .863 | .836 | .862 | .759 | .831 | .739 | .054 | .891 | .881 | .901 | .778 |
| F-Loss [26] | .937 | .033 | .914 | .923 | .854 | .062 | .844 | .852 | .863 | .038 | .864 | .897 | .758 | .690 | .084 | .842 | .829 | .855 | .766 | .798 | .733 | .056 | .884 | .873 | .892 | .794 |
| UAL [6] | .927 | .032 | .925 | .924 | .839 | .059 | .855 | .852 | .842 | .036 | .886 | .901 | .783 | .691 | .080 | .867 | .837 | .868 | .761 | .824 | .734 | .053 | .891 | .873 | .899 | .782 |
| SCLoss | .933 | .032 | .929 | .927 | .845 | .059 | .860 | .858 | .844 | .035 | .891 | .903 | .795 | .701 | .080 | .867 | .850 | .870 | .764 | .833 | .748 | .051 | .895 | .889 | .903 | .784 |

TABLE VIII
SCLOSS COMPATIBILITY STUDY. IN EACH ROW, WE ADD MORE LOSSES TO TWO SOTA SOD MODELS (MENET [24] AND ICON-P [18]). BY ADDING SCLOSS, BOTH MODELS WERE IMPROVED, ACHIEVING NEW SOTA RESULTS.

| | | OMRON | | | | HKU-IS | | | | DUT-TE | | | |
|---|---|---|---|---|---|---|---|---|---|---|---|---|---|
| Model | Loss | $F_\beta^{adp}$ | MAE | $S_m$ | $E_m$ | $F_\beta^{adp}$ | MAE | $S_m$ | $E_m$ | $F_\beta^{adp}$ | MAE | $S_m$ | $E_m$ |
| | + BCE | .778 | .052 | .841 | .868 | .913 | .024 | .927 | .960 | .858 | .031 | .899 | .908 |
| MENet | + $L_g + L_s$ | .790 | .046 | .850 | .881 | .925 | .023 | .927 | .960 | .873 | .028 | .905 | .921 |
| | + SCLoss | .796 | .047 | .856 | .884 | .933 | .022 | .933 | .964 | .877 | .028 | 909 | .921 |
| | + BCE | .785 | .050 | .861 | .881 | .916 | .023 | .933 | .960 | .860 | .026 | .917 | .913 |
| ICON-P | + IoU | .794 | .047 | .865 | .886 | .925 | .022 | .935 | .963 | .868 | .026 | .917 | .919 |
| | + SCLoss | **.808** | **.045** | **.870** | **.893** | **.936** | **.020** | **.940** | **.966** | **.891** | **.024** | **.923** | **.930** |

*C. Impact of α*

Table V assesses the impact of the mutual response and pairwise regularization terms in SCLoss by comparing the performance of the SCLoss-enhanced GCPA and SLSR with $\alpha = 0.5, 1,$ and 2. A lower value of $\alpha$ indicates that the weight will depend more on the mutual response between the pixel pairs and vice versa. The best experimental results were achieved with $\alpha = 1$. Increasing $\alpha$ to 2 lowered the performance. These experiments emphasized the importance of using both terms together to achieve optimal outcomes.

*D. Selection of Pairwise Regularization Term*

In Section III (C), we chose the Gaussian function as the default pairwise regularization term. Table VI compares the performance of other options (among many) for this term.

**Constant regularization**. $f(p_i, p_j) = 1$, is the simplest option to avoid overflow in loss back-propagation, and as expected, it did not result in the best performance.

**Distance regularization**, in the form of the Euclidean distance [57], is defined as:

$$f(p_i, p_j) = e^{(p_i - p_j)^2}. \tag{9}$$

Compared to the Gaussian regularization in (8), this function gives the largest weight when $p_1 = p_2$, which ensures a smooth and polarized prediction [6, 26]. However, it gives less regularization when there is a likely edge prediction. On the other hand, (8) is maximized in cases where both predictions are close to 1, while giving moderate weight to the potential edge predictions. Thus, it reduces the false positives to achieve low MAE and results in predictions with high spatial coherence.

*E. Loss Comparison and Compatibility Study*

**Loss Comparison**. While re-emphasizing that SCLoss can be utilized in combination with previously proposed cost functions, for the sake of completeness, Table VII compares the one-to-one performance of SCLoss and other popular loss functions in SOD and COD. Specifically, we report the performance of GCPANet (SOD) and SLSR (COD) as the backbones using either of the proposed SCLoss, BCE loss, CEL [20], Structure Loss [27], F-loss [26], or UAL [6]. In these one-to-one comparisons, SCLoss produced the overall best results. Compared to F-loss, SCLoss achieved much better results on metrics $MAE$, $S_m$, and E-measures. Compared to all other losses, SCLoss produced the best $MAE$, F-measures, and E-measures.

**Compatibility Study**. Even more encouraging than the one-to-one comparisons, Table VIII shows the compatibility of SCLoss when combined with other loss functions. Following [24], our experiments were based on two SOTA models (ICON [18] and MENet [24]) in SOD by gradually adding different loss functions in these two papers to the models. Noticeably, adding SCLoss to the already complex loss functions improved the models' performance.

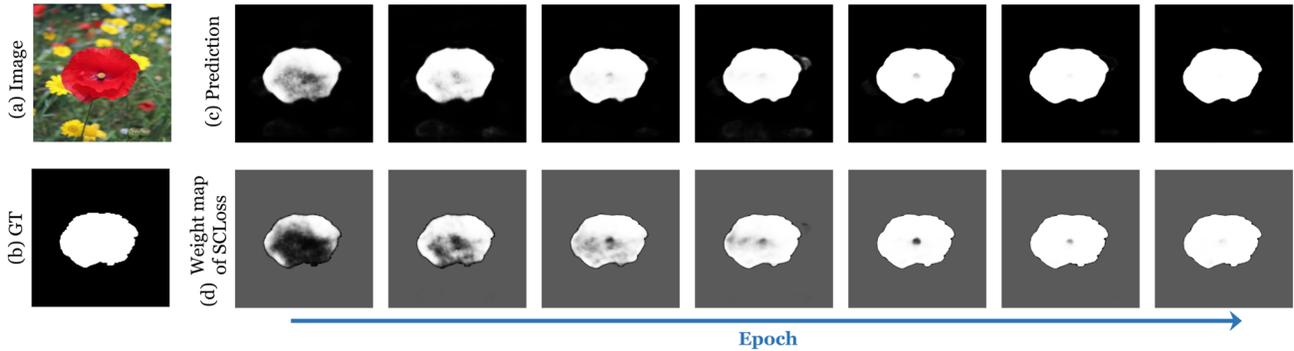

**Fig. 5.** Visualization of the gradual learning process of an example ambiguous region (here: the pistil region) in image (a) with GT (b). Image series (c) are the predictions of the model at different epochs. (d) are the corresponding weight attention maps produced by SCLoss.

*F. Gradual Learning Process of Ambiguous Regions*

Fig. 5 visualizes the learning process of SCLoss when dealing with ambiguous regions. In Fig. 5(c), we show the predictions of a specific image, Fig. 5(a), during the training process. To visualize how SCLoss gave attention to the boundary of the ambiguous regions and gradually learned the ambiguous pixels, Fig. 5(d) shows the corresponding weight attention map of the SCLoss, i.e., one over the denominator of (4). At earlier epochs, when the network could not accurately capture the texture of the flower, the SCLoss emphasized most on the petals. Then, the network gradually gave more attention to the most ambiguous part of the image – the pistil. The network attention started with the intersection of the petal and the pistil, rather than the center of the pistil, to learn the transition between the learned non-ambiguous regions (petal) and the unlearned ambiguous regions (pistil) and gradually moved to the center of the pistil. Eventually, the network produced a perfect detection of the flower.

*G. SCLoss Visualization*

Fig. 6 visually compares the loss outputs of SCLoss with a single-response loss function (BCE). Compared to BCE in Fig. 6(d), which gives high weights to the entire ambiguous regions and dilutes the boundary of the decoy object, the proposed SCLoss in Fig. 6(e) puts more focus on the boundary between the ambiguous and non-ambiguous regions and gradually reduces the focus moving towards the center of the decoy object. These properties enable the network to tackle the ambiguous regions gradually and smoothly by consistently being aware of and learning the transition from non-ambiguous to ambiguous regions.

## VII. DISCUSSION

In this paper, our proposed SCLoss demonstrates superior performance in both SOD and COD domains. Its advantages are evident in several aspects: 1) SCLoss can either replace existing losses or enhance their performance in an add-on manner; 2) SCLoss has different instantiations, and its performance shows stable improvements for different single response terms; 3) SCLoss is highly compatible with existing loss functions. These results demonstrated the feasibility and superiority of integrating the idea of mutual response into traditional objectiveness modeling.

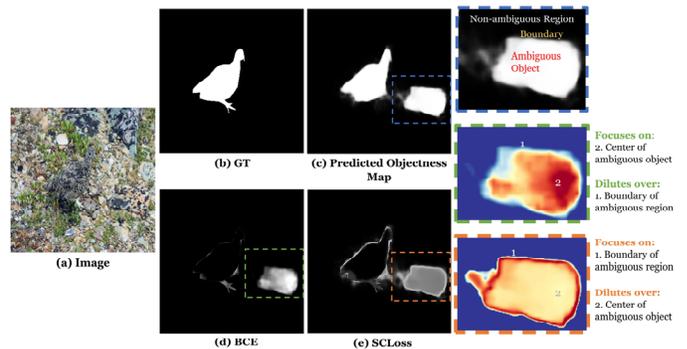

**Fig. 6.** Loss visualization. Given a camouflaged object in (a) and the prediction in (c), the BCE and SCLoss maps are shown in (d) and (e), respectively. Our SCLoss (e) gives a higher weight to the boundary between the non-ambiguous and ambiguous regions, and the weights shrink towards the center of the decoy object to gradually learn the transition from non-ambiguous to ambiguous regions, rather than giving high weight to the center of the ambiguous region in (d).

There is significant room for improvement. First, we used the most common single response loss [25] to construct SCLoss. Considering the compatibility of SCLoss with other loss functions, we expect that more complex instantiations could yield superior outcomes. Second, integrating the proposed loss scheme with the inter-pixel relationship-aware network architecture design [8, 32] is a promising approach to further enhance objectiveness modeling. Third, considering the flexibility and applicability of our framework and formulation, we expect fruitful applications to other similar tasks such as semantic segmentation [70], pose estimation [71], medical image analysis [72], and knowledge distillation [73]. However, it should be noted that for tasks with a significant number of classes, separately modeling mutual response for all classes may increase computational cost in training. For such applications, optimization of algorithm complexity could be a potential research direction.

## VIII. CONCLUSION

This paper introduced a novel loss scheme, SCLoss, for generic object detection. We proposed that utilizing the mutual response between pixels can enhance the spatial coherence of generic object detection. We investigated the utility of the proposed SCLoss by conducting comprehensive experiments on two tasks, SOD and COD. We showed the effectiveness of incorporating SCLoss with the SOTA models as backbones and achieved the new SOTA results on both tasks. The proposed SCLoss was compatible with other loss functions and improved performance when added to existing models. SCLoss was robust with respect to the choice of its individual components, such as the single-response term.